\documentclass[11pt,a4paper]{article}
\usepackage[hyperref]{acl2020}
\usepackage{times}
\usepackage{latexsym}
\usepackage{graphicx}
\usepackage{verbatim}

\usepackage{microtype}

\aclfinalcopy % Uncomment this line for the final submission
\def\aclpaperid{2537} %  Enter the acl Paper ID here

\newcommand{\ttg}{TTG}
\newcommand{\tgt}{TGT}
\newcommand{\trg}{X}
\newcommand{\plugs}{PLuGS}
\newcommand{\sxs}{SxS}  % Side-by-side
\newcommand{\oid}{OID1k}  % OID-1k Images used for human eval
\newcommand{\ok}{Accept}  % Caption is Mediocre, Good or Excellent (i.e non-bad)

\title{{C}ross-modal {L}anguage {G}eneration using {P}ivot {S}tabilization for {W}eb-scale {L}anguage {C}overage}

\author{Ashish V. Thapliyal \\
  Google Research \\
  \texttt{asht@google.com} \\\And
  Radu Soricut \\
  Google Research \\
  \texttt{rsoricut@google.com} \\}

\date{}

\begin{document}

\maketitle

\begin{abstract}
Cross-modal language generation tasks such as image captioning
are directly hurt in their ability to support non-English languages
by the trend of data-hungry models combined with the lack of non-English annotations.
We investigate potential solutions for combining existing language-generation annotations
in English with translation capabilities in order to create solutions at
web-scale in both domain and language coverage.
We describe an approach called Pivot-Language Generation Stabilization (PLuGS),
which leverages directly at training time both existing English annotations
(gold data) as well as their machine-translated versions (silver data);
at run-time, it generates first an English caption and
then a corresponding target-language caption.
We show that PLuGS models outperform other candidate solutions in
evaluations performed over 5 different target languages,
under a large-domain testset using images from the Open Images dataset.
Furthermore, we find an interesting effect where the English captions
generated by the PLuGS models are better than
the captions generated by the original, monolingual English model.

\end{abstract}

\section{Introduction}
\label{sec:intro}
% The current direction in terms of training machine-learned models that generate
% language has the potential to widen an already-large gap between the status
% of English compared to other languages: data-hungry state-of-the-art neural models
% require large amounts of labeled data in order to adequately learn to generate outputs
% in language X, whereas annotated data for generation in language X is extremely scarce
% for languages X different than English.
Data hungry state-of-the-art neural models for language generation have the undesired potential
to widen the quality gap between English and non-English languages, given the scarcity of non-English labeled data.
One notable exception is machine translation, which benefits from large
amounts of bilingually or multilingually annotated data.
But cross-modal language generation tasks, such as automatic image captioning, tend
to be directly hurt by this trend: existing datasets such as Flickr~\cite{flickr30k}, MSCOCO~\cite{coco},
and Conceptual Captions~\cite{sharma2018conceptual} have extensive labeled data for English, but
labeled data is extremely scarce in other languages~\cite{ElliottFSS16}
(at 2 orders of magnitude less for a couple of languages, and none for the rest).

In this paper, we conduct a study aimed at answering the following question:
given a large annotated web-scale dataset such as Conceptual Captions~\cite{sharma2018conceptual}
in one language, and a baseline machine translation system,
what is the optimal way to scale a cross-modality language generation system to new languages at web-scale?

\begin{figure*}[t!]
  \begin{center}
   \includegraphics[width=1.0\linewidth]{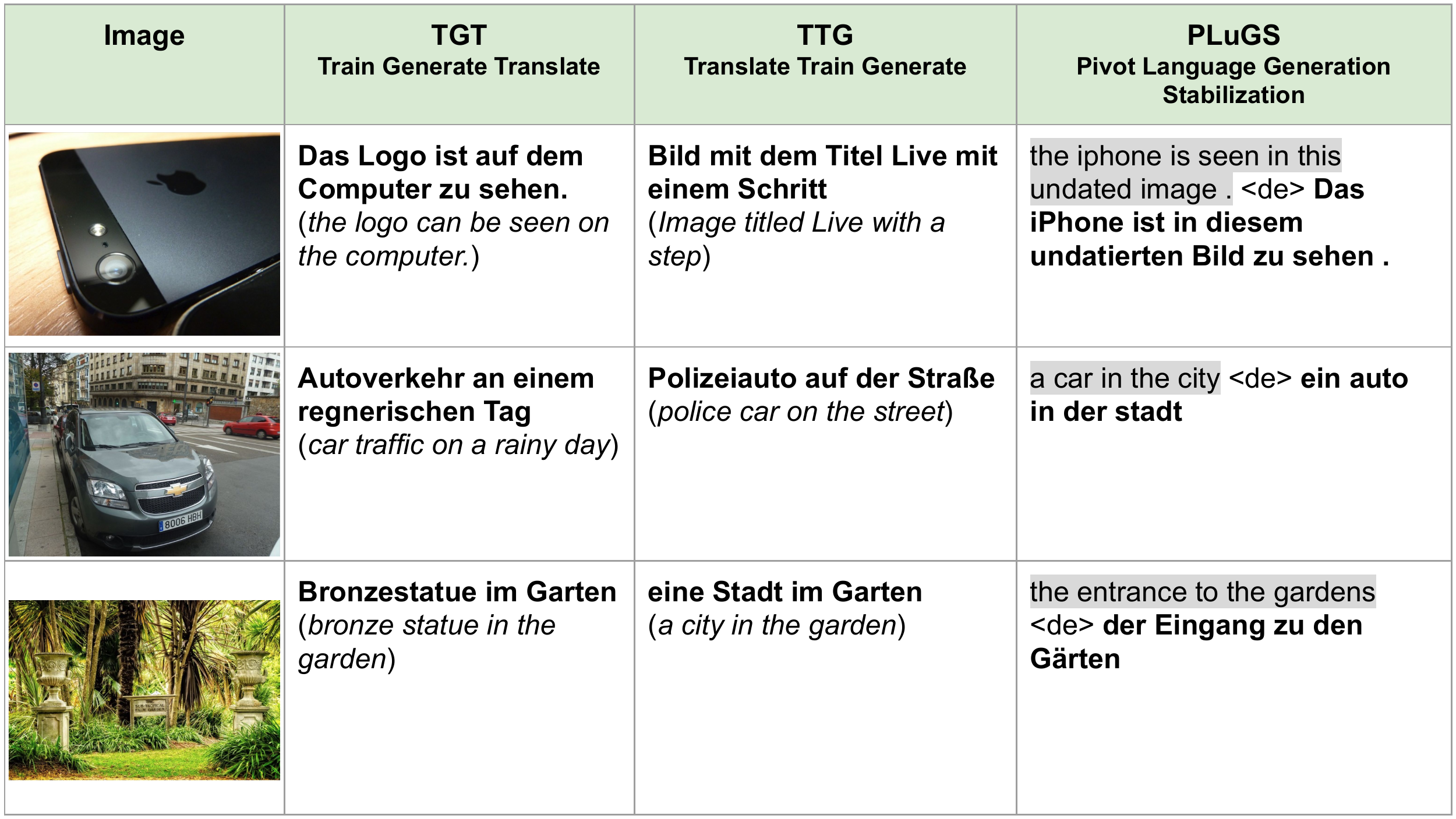}
   \caption{\label{fig-first_ex}Examples of captions produced in German by Train-Generate-Translate (\tgt), Translate-Train-Generate (\ttg), and Pivot Language Generation Stabilization (\plugs) approaches.
   Captions are shown in bold font. For \tgt~and \ttg~outputs, we show the English translation in parenthesis beside the caption. For the \plugs~outputs
   we mark the Stabilizer in the output using a light gray background. We do not explicitly show a translation for \plugs~outputs since the Stabilizer
   is already a translation. }
  \end{center}
\end{figure*}

We focus our study on the task of automatic image captioning, as a representative for
cross-modal language generation where back-and-forth consistency
cannot be leveraged in a straightforward manner
\footnote{We chose to focus on the cross-modality version of this problem because for the text-only
modality the problem is less severe (due to existing parallel data) and also
more studied~\cite{artetxe2018unsupervised}, as it is amenable to exploiting back-and-forth
consistency as a powerful learning signal.}.
In this framework, we proceed to test several possible solutions, as follows:
(a) leverage existing English (En) image captioning datasets to train a model that generates
En captions, which are then translated into a target language \trg; we call this approach Train-Generate-Translate (\tgt);
(b) leverage existing En captioning datasets and translation capabilities to first
translate the data into the target language \trg, and then train a model that generates \trg~-language captions;
we call this approach Translate-Train-Generate (\ttg);
(c) stabilize the \ttg~approach by directly using the En gold data along with the translated training data in the \trg~language (silver data)
to train a model that first generates En captions (conditioned on the image),
and then generates \trg~-language captions (conditioned on the image and the generated En caption);
this approach has En acting as a pivot language between the input modality and
the \trg~-language output text, stabilizing against and reducing potential translation noise.
We call the latter the Pivot-Language Generation Stabilization (\plugs) approach.
Examples of outputs produced by these three solutions are shown in Fig.~\ref{fig-first_ex}.

We perform extensive evaluations across five different languages (French, Italian, German, Spanish, Hindi) to compare these three approaches.
The results indicate that the bilingual \plugs~models consistently perform the best in terms of captioning accuracy.
Since there is very little support in the literature regarding the ability of
standard evaluation metrics like BLEU~\cite{papineni-etal:2002}, ROUGE~\cite{lin2004rouge}, METEOR~\cite{meteor},
CIDEr~\cite{cider}, and SPICE~\cite{spice} to accurately measure
captioning accuracy for non-English languages, our evaluations are done using fine-grained, side-by-side human evaluations
using paid raters; we explain the evaluation protocol in detail in Sec.~\ref{sec:eval}.

Besides the evaluations on bilingual \plugs~models, we also train and evaluate a multilingual \plugs~model, in which
all five non-English languages considered are supported through a single model capable of generating outputs in all 5 languages.
The results indicate that similar languages are reinforcing each other in the common representation space,
showing quantitative gains for the Romance languages involved in our experiments.
A related but perhaps less expected result is that the English captions generated by \plugs~models (what we call the Stablizer outputs) are better,
as measured using side-by-side human evaluations, than captions generated by the original, monolingual English model.

There is a final additional advantage to having \plugs~models as a solution:
in real-world applications of image captioning, quality estimation of the resulting
captions is an important component that has recently received
attention~\cite{levinboim2019quality-arxiv}.
Again, labeled data for quality-estimation (QE) is only available for
English\footnote{https://github.com/google-research-datasets/Image-Caption-Quality-Dataset},
and generating it separately for other languages of interest is expensive, time-consuming, and scales poorly.
The \tgt~approach could directly apply a QE model at run-time on the En caption, but the subsequent translation step would need to be
perfect in order not to ruin the predicted quality score.
The \ttg~approach cannot make use at run-time of an En QE model without translating the caption back to
English and thus again requiring perfect translation in order not to ruin the predicted quality score.
In contrast, the \plugs~approach appears to be best suited for leveraging an existing En QE model, due to the availability of the
generated bilingual output that tends to maintain consistency between the generated EN- \& X-language outputs, with respect to accuracy;
therefore, directly applying an English QE model appears to be the most appropriate scalable solution.

\section{Related Work}
\label{sec:related}
There is a large body of work in automatic image captioning for English, starting with
early work~\citep{hodosh13framing,donahue2014long,karpathy2014deep,kiros2014unifying,xu15show}
based on
data offered by manually annotated datasets such as Flickr30K~\citep{young2014from} and MS-COCO~\cite{coco},
and more recently with work using Transformer-based
models~\citep{sharma2018conceptual,zhao2019informative,changpinyo2019decoupled}
based on the web-scale Conceptual Captions dataset~\citep{sharma2018conceptual}.

Generating image captions in languages other than English
has been explored in the context
of the WMT 2017-2018 multimodal translation sub-task on multilingual caption
generation~\cite{ElliottFBBS17}.
The goal of the task is to generate
image captions in German and French, using a small training corpus with images
and captions available in English, German and French (based on Flickr30K).
In the context of that work, we use the results reported in \cite{CaglayanMSB19} to quantitatively compare it against our approach.

Another relevant connection is with the work in \cite{Jaffe17}, which explores
several LSTM-based encoder-decoder models that
generate captions in different languages.
%while ignoring the English caption output for evaluation.
The model most similar to our work is their Dual Attention model,
which first generates an English caption, then an LSTM with
attention over the image and the generated English caption produces a
German caption.
Their quantitative evaluations do not find any additional benefits for this approach.

Our work is related to this idea, but there are key technical differences.
In the \plugs~approach, we train an end-to-end model based on a Transformer~\cite{vaswani2017attention}
decoder that exploits the generated English-prefix via the self-attention
mechanism to learn to predict the non-English target caption, conditioned on
the English tokens at multiple levels through the decoder stack.
Moreover, we approach this study as the search for a solution for web-scale multi-language image
captioning: we employ the web-sized Conceptual Captions dataset for training, and
consider the effects of using captions across multiple languages, as well as multi-language/single-model setups.

\section{Model Architecture}
\label{sec:model}
\begin{figure*}[h!]
  \begin{center}
   \includegraphics[width=1.0\linewidth]{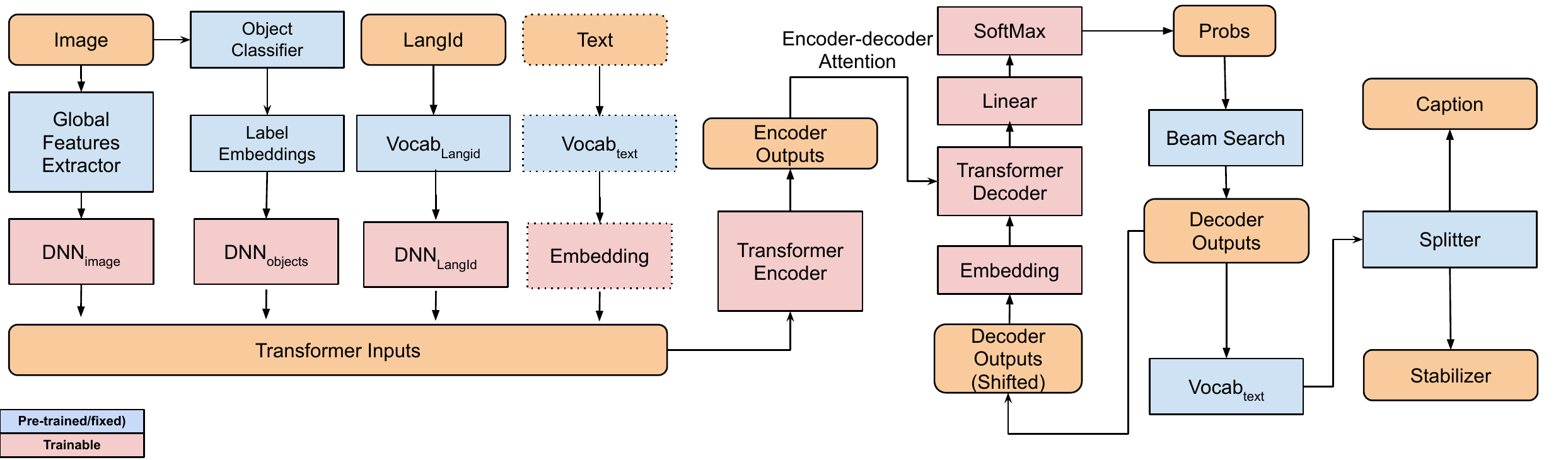}
   \caption{\label{fig:model} The Transformer based \plugs~model. The text on the input side is used
   for the translation and multi-modal translation experiments with the Multi30K dataset. For image captioning, no
   text input is provided.}
  \end{center}
\end{figure*}
We model the output caption using a sequence-generation approach based on Transformer Networks \cite{vaswani2017attention}.
The output is the sequence of sub-tokens comprising the target caption.
As shown in Fig.~\ref{fig:model}, the input sequence is obtained by concatenating the following features.

%We use a Transformer\cite{vaswani2017attention} based model and train it to minimize the ground-truth text perplexity.

\paragraph{Global Image Embedding:}
We use a global image representation using the Graph-RISE model~\cite{juan2019graphrise},
a ResNet-101 model~\cite{resnet2016} trained for image classification at ultra-fine granularity levels.
This model produces a compact image embedding $i$ of dimension $D_i=64$.
This embedding is projected to match Transformer dimensions (set to 512 in most of our experiments)
by a 2 layer DNN with linear activation
and fed as the first element in the sequence of inputs to
the encoder.

\paragraph{Object Labels Embeddings:}
Detecting the presence of certain objects in the image
(e.g. ``woman'', ``flag'', ``laptop'') can help
generate more accurate captions, since a good caption should mention the more salient objects.
The object labels are generated by an object detection model which is run over
the entire image.
The output labels are then converted to vectors using word embeddings to obtain
what we call object-label embeddings.

More precisely, we detect object labels over the entire image using a ResNet-101 object-detection
classifier trained on the JFT dataset~\cite{hinton2015distilling}.
The classifier produces a list of detected object-label identifiers,
sorted in decreasing order by the classifier's confidence score; we use the first sixteen of these identifiers.
The identifiers are then mapped to embeddings $o_j$ using an object-label embedding layer which is pre-trained
to predict label co-occurrences in web documents, using a word2vec approach~\cite{mikolov-et-al:2013a}.
The resulting sequence of embeddings is denoted $O=(o_1, \ldots, o_{|O|})$,
where each $o_j$ has dimension $D_o=256$. Each member of this sequence of embeddings is
projected to match Transformer dimensions by a 2 layer DNN with linear activation. This
sequence of projected object-label embeddings is fed to the encoder together with the global image embedding.

\paragraph{LangId Embeddings:}
When training language-aware models, we add as input the language of the target sequence.
We specify the language using a language identifier string such as $en$ for English, $de$ for German, etc.
We call this the LangId of the target sequence or target LangId in short.
Given the target LangId, we encode it using a
LangId vocabulary, project it to match Transformer dimensions with a 2 layer DNN,
then append it to the encoder input sequence.

\paragraph{Text Embeddings:}
All text (input or output) is encoded using byte-pair encoding~\cite{sennrich-etal:16} with a shared source-target
vocabulary of about 4000 tokens, then embedded as described in \cite{vaswani2017attention},
resulting in a sequence of text embeddings.
The embeddings dimensions are chosen to match the Transformer dimensions.
When performing the
translation (MT) and multimodal translation (MMT) experiments in Sec.~\ref{sec:multi30k},
the sequence of source text embeddings are fed to the encoder after the LangId embedding.
Additionally, we reserve a token-id in the text vocabulary for each language
(e.g. $\langle de\rangle$ for German) for use as a separator in the
\plugs~model output and also have a separate start-of-sequence token for each language.

\paragraph{Decoding:} We decode with beam search with beam width 5.

\begin{figure}[t!]
  \begin{center}
   \includegraphics[width=1.0\linewidth]{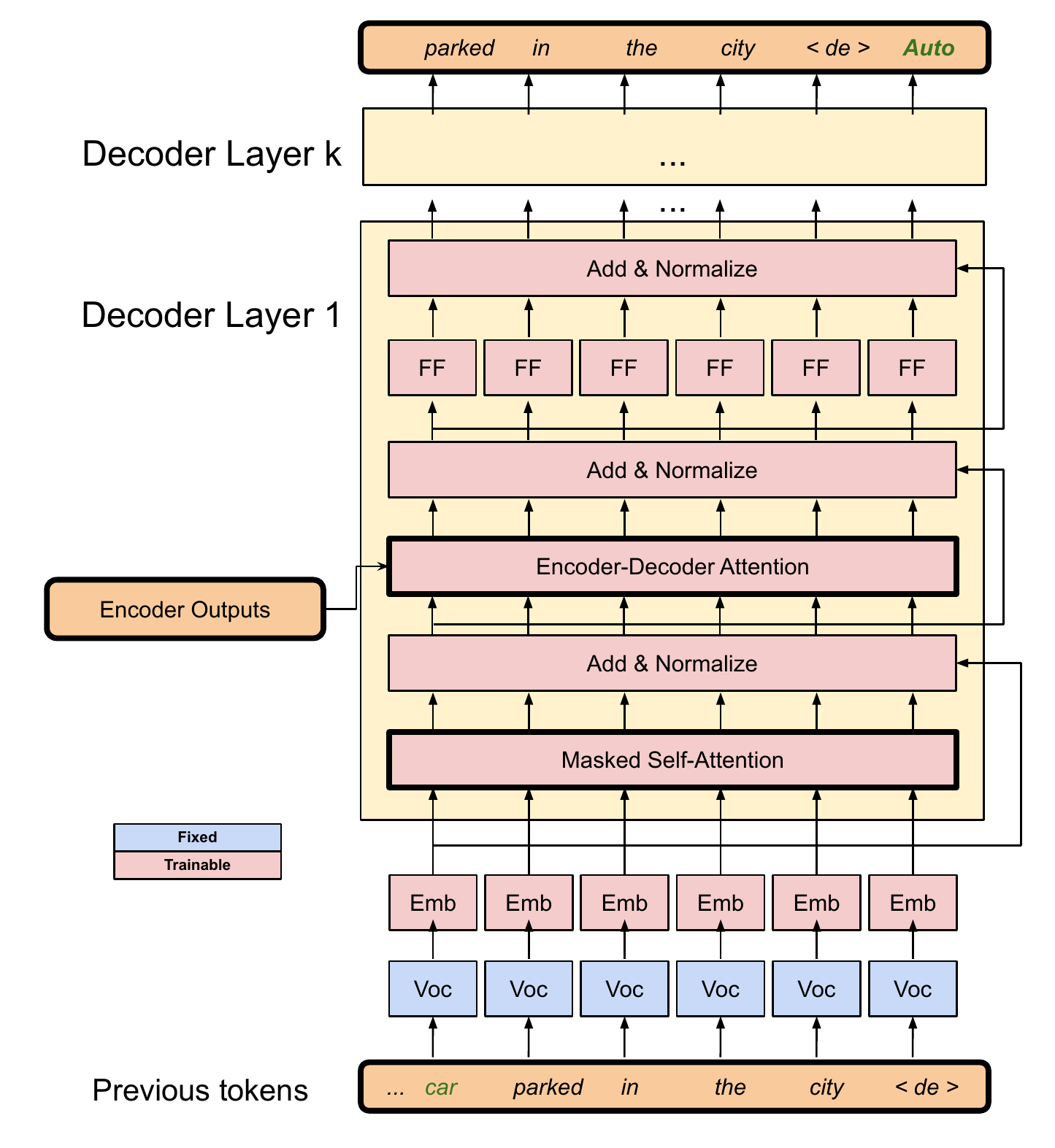}
   \caption{\label{fig:depend} Caption's dependence on the Stabilizer.
   The target-language caption is conditioned on the Stabilizer through the Masked Self-Attention in the decoder,
   and on the input image through the Encoder-Decoder attention
   that attends to the outputs of the last encoder layer.
   Note that in this figure, FF stands for the feed forward network, Voc stands for the (fixed) text vocab, and Emb stands for the (trainable) text embeddings.}
  \end{center}
\end{figure}

\paragraph{PLuGS:} For PLuGS~models, in addition to the target caption we require
the model to generate a pivot-language (En) caption which we call the Stabilizer.
Specifically, we train the model over target sequences of the form Stabilizer + $\langle$separator$\rangle$ + Caption.

We use $\langle$\$LangId$\rangle$ as the separator (i.e., for German captions we
use $\langle de \rangle$ as the separator).
This approach has the advantage that it can be applied to multilingual models as well.
We subsequently split the model output based on the separator to obtain two strings: the Stabilizer and the Caption.

Note an important technical advantage here: as shown in Fig.~\ref{fig:depend},
after initially generating the Stabilizer output,
the Transformer decoder is capable of exploiting it directly via the self-attention mechanism,
and learn to predict the non-English Caption tokens conditioned (via teacher-forcing) on
the gold-data English tokens at multiple levels through the decoder stack, in addition to the cross-attention
mechanism attending to the inputs.
As our results indicate, the models are capable of maintaining this advantage at run-time as well, when
auto-regressive decoding is performed.

\section{Datasets}
\label{sec:datasets}
We perform our experiments using two different benchmarks.
We use the Multi30K~\cite{ElliottFSS16} dataset in order to compare the effect of
the \plugs~model using a resource that has been widely used in the community.
We focus on Task 1 for French from~\cite{CaglayanMSB19}, generating a translation in French based on an image and an English caption as input.
The training set consists of images from the Flickr30K train and validation splits, along with the corresponding French captions.
The validation split consists of test2016 images and captions, and the test split consists of the test2017 images and captions.

For the core results in this paper, we use the Conceptual Captions dataset~\cite{sharma2018conceptual}
as our English-annotated generation labels, in order to capture web-scale phenomena related to image captioning.
In addition, we use Google Translate as the translation engine
(both for the run-time translations needed for the \tgt~approach and the training-time translations needed for the \ttg~and \plugs~approaches),
targeting French, Italian, German, Spanish, and Hindi as target languages.
We use the standard training and validation splits from Conceptual Captions for developing our models.
We report the results using
%1,000 images from
%the publicly-available\footnote{http://www.conceptualcaptions.com/winners-and-data}
% test set for the image-caption competition for the Conceptual Caption Workshop at the
%2019 Conference on Computer Vision and Pattern Recognition, consisting of a
a set of 1,000 randomly samples images
from the Open Images Dataset~\cite{oidv4}.
We refer to this test set as \oid~when reporting our results.

\begin{figure*}[h!]
  \begin{center}
   \includegraphics[width=1.0\linewidth]{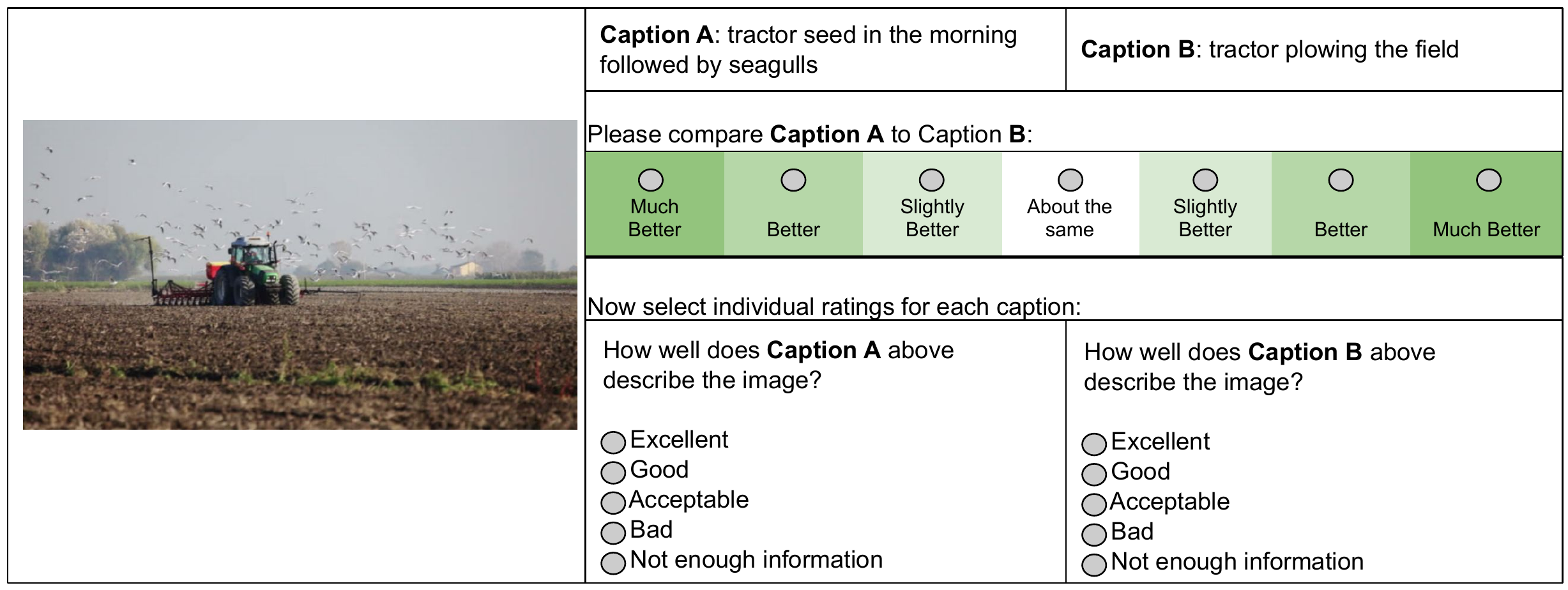}
   \caption{\label{fig:sxs}Side-by-side human evaluation of two image captions.
   The same template is used for evaluating English as well as the 5 languages targeted.}
  \end{center}
\end{figure*}

\section{Evaluation}
\label{sec:eval}
In the experiments done using the Multi30K dataset, we are reporting results using the METEOR~\cite{meteor} metric, in line with previous work.
For the experiments performed using the Conceptual Captions dataset, we have found that automated evaluation metrics for image captioning such as
 BLEU~\cite{papineni-etal:2002}, ROUGE~\cite{lin2004rouge}, CIDEr~\cite{cider}, and SPICE~\cite{spice} cannot accurately measure
captioning accuracy for non-English languages.
However, we are reporting CIDEr numbers as a point of comparison, and contrast these numbers with human evaluation results.
We describe the human evaluation framework we use next.

\subsection{Human Side-by-Side Evaluation}

We perform side-by-side human evaluation for comparing model outputs.
To compare two image captioning models $A$ (baseline) vs $B$, we generate captions for these images with each model and ask
human raters to compare them.
As illustrated in Fig.~\ref{fig:sxs}, the raters are shown the image with the two captions randomly placed to the left vs. right,
and are asked to compare the captions on a side-by-side rating scale.
In addition, they are asked to also provide an absolute rating for each caption.
The absolute rating provides a cross-check on the comparison.
Each image and associated captions are rated by three raters in our experiments.

We calculate the following statistics using the resulting side-by-side rating comparisons:\\
$Wins$: Percent of images where majority of raters (i.e. 2 out of 3) marked Caption B as better (after derandomization).\\
$Losses$: Percent of images where majority of raters marked Caption A as better.\\
$Gain_{sxs} = Wins - Losses$

We also calculate the following statistics using the resulting absolute ratings:\\
$A_{\ok}$ = Percent of images where majority of raters mark caption A as Acceptable, Good, or Excellent. \\
$B_{\ok}$ = Percent of images where majority of raters mark caption B as Acceptable, Good, or Excellent. \\
$Gain_{\ok} = B_{\ok} - A_{\ok}$

The advantages of the $Gain_{sxs}$ and $Gain_{\ok}$ metrics is that they are intuitive, i.e., they measure the absolute increase
in accuracy between the two experimental conditions\footnote{Inter-rater agreement
analysis shows that for each evaluation comparing two models, two of the
three raters agree on $Win/Loss/Same$ for $90\%$ to $95\%$ of the items.
Further, for more than $98\%$ of the items using the difference between the
absolute ratings gives the same $Win/Loss/Same$ values as obtained from the side-by-side ratings.
Also, for $80\%$ to $85\%$ of the absolute ratings, two of the three raters agree on the rating.}

\subsection{Training Details}
\label{sec:train-details}
\paragraph{Multi30K:}
For the experiments using this dataset, we use a Transformer Network~\cite{vaswani2017attention}
with 3 encoder and 3 decoder layers, 8 heads, and model dimension 512.
We use the Adam optimizer~\cite{Kingma2015AdamAM}, and do a hyperparameter search over
learning rates $\{3e^{-4}, e^{-4}, 3e^{-5}, e^{-5}\}$ with linear warmup over 16000 steps
followed by exponential decay over $\{50k, 100k\}$ steps. We use $5e^{-6}$ as the
weight for $L_2$ regularization. We train with a batch size of 1024, using a dropout of 0.3,
on 8 TPU~\cite{YouZHDK19} cores.

\paragraph{Conceptual Captions:}
For all except large multilingual models, we use a vanilla Transformer
with 6 encoder and decoder layers, 8 heads, and model dimension 512.
We use the SGD optimizer, and do a hyperparameter search over
learning rates $\{0.12, 0.15, 0.18, 0.21, 0.24\}$ with linear warmup over 16000 steps
followed by exponential decay over $\{350k, 450k\}$ steps.
For multilingual models, we also use linear warmup over 80000 steps.
We use $1e^{-5}$ as the weight for $L_2$ regularization.
We train with a batch size of 4096, using a dropout of 0.3 on 32 TPU~\cite{YouZHDK19} cores.

For large multilingual models, we use a Transformer with 10 encoder and decoder layers,
12 heads, and model dimension 768\footnote{Dimension chosen so that we maintain 64 dimensions per head.}
We also use a smaller learning rate of 0.09.

\begin{table}[bh]
\centering
\begin{tabular}{llll}
\hline \textbf{Task} & Baseline  & non-\plugs        & \plugs \\ \hline
MT                     & 70.6            & 66.6             &          67.7            \\
MMT                    & 70.9            & 64.7             &          65.6            \\
IC-D$_4$                 & 32.3            & 30.6             &          32.8            \\
\hline
\end{tabular}
\caption{\label{tab:multi30k} Multi30K test set METEOR scores for Translation (MT),
Multi Modal Translation (MMT), and Image Captioning (IC-D$_4$).
The baseline is from task 1 of \cite{CaglayanMSB19}. }
\end{table}

\begin{table*}[th]
\centering
\begin{tabular}{llllllllll}
\hline
\textbf{Lang} & \textbf{$Wins$} & \textbf{$Losses$} & \textbf{$Gain_{sxs}$} & \textrm{\plugs$_{\ok}$} & \textrm{\tgt$_{\ok}$} & \textbf{$Gain_{\ok}$} \\ \hline
          Fr  &           22.8   &            19.4   &           3.4           &          68.7  &            66.5 &            2.2           \\
          It  &           22.5   &            18.3   &           4.2           &          52.1  &            49.9 &            2.2           \\
          De  &           22.6   &            19.1   &           3.5           &          69.2  &            67.7 &            1.5           \\
          Es  &           27.0   &            22.1   &           4.9           &          58.8  &            56.9 &            1.9           \\
          Hi  &           26.8   &            23.8   &           3.0           &          78.6  &            75.9 &            2.7           \\
\hline
              & \textbf{$Wins$} & \textbf{$Losses$} & \textbf{$Gain_{sxs}$} & \textrm{\plugs$_{\ok}$} & \textrm{\ttg$_{\ok}$} & \textbf{$Gain_{\ok}$} \\ \hline
          Fr  &           18.2   &            17.3   &           0.9           &          66.2  &            64.2 &            2.0          \\
          It  &           23.7   &            20.8   &           2.9           &          55.1  &            52.2 &            2.9          \\
          De  &           21.9   &            19.6   &           2.3           &          64.3  &            63.0 &            1.3          \\
          Es  &           24.9   &            23.8   &           1.1           &          57.7  &            56.8 &            0.9          \\
          Hi  &           27.4   &            25.5   &           1.9           &          71.3  &            69.6 &            1.7          \\
\hline
\end{tabular}
\caption{\label{tab:tgt-ttg_vs_plugs} SxS performance of \plugs~vs. \tgt~models (upper half) and \plugs~vs.
\ttg~ models (lower half), across five target languages on \oid.
The \plugs~models perform better on both $Gain_{\sxs}$ and $Gain_{\ok}$ metrics, for all five languages.}
\end{table*}

\begin{table*}
\centering
\begin{tabular}{llllllllll}
\hline
\textbf{Lang} & \tgt    & \ttg     & \plugs   & \plugs-\tgt & \plugs-\ttg \\ \hline
          Fr  &  0.7890 &  0.7932  & 0.7820   & -0.0070     & -0.0112       \\
          It  &  0.7729 &  0.7760  & 0.7813   &  0.0084     &  0.0053       \\
          De  &  0.6220 &  0.6079  & 0.6170   &  0.0050     &  0.0091       \\
          Es  &  0.8042 &  0.7907  & 0.7854   & -0.0188     & -0.0053       \\
          Hi  &  0.7026 &  0.7149  & 0.7155   &  0.0129     &  0.0006       \\
\hline
\end{tabular}
\caption{\label{tab:cider_tgt_ttg_plugs} CIDEr scores on CC-1.1 validation set for \plugs, \tgt, and \ttg~models for five languages.}
\end{table*}

\section{Experiments and Results}
\label{sec:results}

\subsection{Multi30K}
\label{sec:multi30k}
In order to compare our work to related work we train our models on the Multi30K dataset
and compared our results to the results in \cite{CaglayanMSB19}.
We focus on Task 1: generate a French translation based on an image and English caption as input.
Table~\ref{tab:multi30k} shows the results on the Multi30K dataset for Multimodal Translation.
Note that since~\cite{CaglayanMSB19} does not show numbers for the pure (no caption input) image captioning task, we
show numbers for the $D_4$ condition, where only the first 4 tokens of the English caption
are provided as input to the image captioning model.

We see that the \plugs~model is able to produce numbers for MT and MMT that are close to the
baseline, even thought it is just an image captioning model augmented to handle these tasks.
For the $D_4$ task, which is the closest to image captioning, the \plugs~model shows improvement over the baseline.
Furthermore, the results contain preliminary indications that the
\plugs~approach produces better results compared to the non-\plugs~ approach (+2.2 METEOR).

\subsection{Conceptual Captions}

In this section, we evaluate the performance of models trained using Conceptual Captions, as detailed in Sec.~\ref{sec:datasets}.
Table~\ref{tab:tgt-ttg_vs_plugs} presents the results on the \oid~testset for the \sxs~human evaluations between the \tgt~and \plugs~models (upper half),
and between the \ttg~and \plugs~models (lower half).
The results show that, for all five languages, the \plugs~model captions are consistently superior to the \tgt~captions on both $Gain_{\sxs}$ and $Gain_{\ok}$ metrics.
The $Gain_{\sxs}$ are between 3\% and 5\% absolute percentages between \tgt~and \plugs~models,
and 1\% and 3\% absolute percentages between \ttg~and \plugs~models, with similar trends for the $Gain_{\ok}$ metric.

Table~\ref{tab:cider_tgt_ttg_plugs} presents the CIDEr scores on the validation set of the Conceptual Captions v1.1 (CC-1.1).
The CIDEr metric fails to capture any meaningful correlation between its scores and the results of the SxS human evaluations.

\begin{table*}[ht]
\centering
\begin{tabular}{llllll}
\hline \textbf{Lang} &         \ttg          &       \plugs-2L          &          \ttg-5L    &          \ttg$_\textrm{large}$-5L &   \plugs-5L \\ \hline
                   Fr    &    0.7932        &       0.7820             &          0.6834    &           0.7064         &          0.7264            \\
                   It    &    0.7760        &       0.7813             &          0.6538    &           0.6885         &          0.6978            \\
                   De    &    0.6079        &       0.6170             &          0.4992    &           0.5367         &          0.5503            \\
                   Es    &    0.7907        &       0.7854             &          0.7093    &           0.7203         &          0.7284            \\
                   Hi    &    0.7149        &       0.7155             &          0.5891    &           0.6201         &          0.6641            \\
\hline
\end{tabular}
\caption{\label{tab:cider} CIDEr scores on CC-1.1 validation set for bilingual and multilingual models.}
\end{table*}

\begin{table*}
\centering
\begin{tabular}{llllllllll}
\textbf{Lang} & \textbf{$Wins$} & \textbf{$Losses$} & \textbf{$Gain_{sxs}$} & \textbf{$B_{\ok}$} & \textbf{$A_{\ok}$} & \textbf{$Gain_{\ok}$}  \\ \hline
          Fr  &          21.3    &           18.3    &           3.0           &         69.8   &           68.7  &            1.1         \\
          It  &          22.2    &           18.2    &           4.0           &         56.4   &           55.5  &            0.9         \\
          Hi  &          26.8    &           27.0    &          -0.2           &         75.6   &           79.5  &           -3.9         \\
\hline
\end{tabular}
\caption{\label{tab:sxs_multilang} SxS performance of \plugs-5L vs. \plugs-2L models for three languages.}
\end{table*}

\begin{table*}[ht]
\centering
\begin{tabular}{llllllllll}
\textbf{Model} & \textbf{$Wins$} & \textbf{$Losses$} & \textbf{$Gain_{sxs}$} & \textbf{$B_{\ok}$} & \textbf{$A_{\ok}$} & \textbf{$Gain_{\ok}$}\\ \hline
    \plugs-Fr  &           26.9   &            21.8   &           5.1        &          70.4    &          67.0    &            3.4         \\
    \plugs-De  &           26.6   &            21.3   &           5.3        &          70.4    &          69.7    &            0.7         \\
    \plugs-Es  &           28.0   &            21.8   &           6.2        &          69.7    &          67.8    &            1.9         \\
\hline
\end{tabular}
\caption{\label{tab:stab} Performance of Stabilizers used as captions from
\plugs~models for three languages vs the captions produced by the baseline English model.
The \plugs~Stabilizer outputs are better captions across all three languages.}
\end{table*}

\begin{table}[ht]
\centering
\begin{tabular}{llll}
\textbf{Model} &     \plugs  &          Baseline  &         Diff       \\ \hline
    \plugs-Fr  &     0.8663  &           0.8772   &          -0.0139       \\
    \plugs-De  &     0.8680  &           0.8772   &          -0.0092       \\
    \plugs-Es  &     0.8590  &           0.8772   &          -0.0182       \\
\hline
\end{tabular}
\caption{\label{tab:cider-stab} CIDEr scores on CC-1.1 validation set for Baseline and \plugs-Stabilizer outputs (English captions).}
\end{table}

\subsection{Multilingual Models}
We further explore the hypothesis that
adding more languages inside one single model may perform even better, as a result of
both translation noise canceling out and the languages reinforcing each other in a common representation space.
In this vein, we rename the bilingual version as \plugs-2L, and train several additional models:
a \ttg-5L model, which uses a LangId token as input and uses for training all translated captions for all five languages and English;
a \ttg$_\textrm{large}$-5L model, for which we simply increased the capacity of the Transformer network (see Sec.~\ref{sec:train-details});
and a \plugs-5L model, which is trained using groundtruth labels that are concatenations (using the LangId token as separator) between
golden groundtruth En labels and their translated versions, for all five target languages.

Results using CIDEr are shown in Table \ref{tab:cider}.
Across all languages, the \ttg-5L models show a large gap in the CIDEr scores as compared to the \ttg~monolingual models.
Using more capacity in the \ttg$_\textrm{large}$-5L model closes the gap only slightly.
However, the effect of using pivot-language stabilizers tends to be consistently larger, in terms of CIDEr improvements,
than the ones obtained by increasing the model capacity.

To accurately evaluate the impact of multi-linguality,
we also perform SxS evaluations between the \plugs-2L (as the base condition) vs. \plugs-5L (as the test condition) models, over three languages (French, German, and Hindi).
As shown in Table \ref{tab:sxs_multilang}, the \plugs-5L model performs better on French and Italian (3\% and 4\% better on $Gain_{sxs}$), while performing worse on Hindi compared
to the bilingual \plugs~Hindi model (-0.2\% on $Gain_{sxs}$, -3.9\% on $Gain_{\ok}$).
The results are encouraging, and indeed support the hypothesis that similar languages are reinforcing each other in the common representation space,
explaining the gain observed for the Romance languages and the detrimental impact on Hindi.

We also note here that the human evaluation results, except for Hindi, come in direct contradiction to the CIDEr metric results, which indicate a large performance hit for \plugs-5L vs. \plugs-2L, across all languages.
This reflects again the extreme care needed when judging the outcome of such experiments based on the existing automatic metrics.

\subsection{Stabilizers Used as English Captions}
\begin{table}[ht]
\centering
\begin{tabular}{llll}
\hline \textbf{Model} & \multicolumn{3}{c}{Spearman $\rho$} \\
                     & \tgt              & \ttg              & \plugs            \\ \hline
          \plugs-Fr  &       0.3017      &          0.3318   &         {\bf 0.5982}    \\
          \plugs-De  &       0.3246      &          0.2900   &         {\bf 0.5862}    \\
          \plugs-Es  &       0.2928      &          0.3201   &         {\bf 0.5566}    \\
\hline
\end{tabular}
\caption{\label{tab:corr} Spearman correlation of Stabilizer vs \tgt, \ttg~and
\plugs~Captions across three languages.}
\end{table}

\begin{table}[ht]
\centering
\begin{tabular}{llllll}
\hline  & Fr     &  It    & De     & Es     & Hi     \\ \hline
          BLEU     & 93.26  & 92.86  & 88.21  & 93.88  & 88.15  \\
\hline
\end{tabular}
\caption{\label{tab:bleu} The BLEU-4 score of the translation of the stabilizer against the caption treated as the reference.}
\end{table}
As already mentioned, the \plugs~models generate outputs of the form Stabilizer + $\langle$LangId$\rangle$ + Caption.
We therefore ask the following question: how does the quality of the Stabilizer output compare to the quality of
captions produced by the baseline English model
(that is, the same model whose captions are translated to the target languages in the \tgt~approach)?

We perform SxS human evaluations over Stabilizer captions (English) for three different \plugs-2L models (trained for French, German, and Spanish).
As shown in Table~\ref{tab:stab}, the somewhat unexpected answer is that these Stabilizer outputs are consistently
better, as English captions, compared to the ones produced by the original monolingual English captioning model.
The $Gain_{sxs}$ are between 5\% and 6\% absolute percentage improvements, while $Gain_{\ok}$ also improves up to 3.4\% absolute for the \plugs-Fr model.

We again note that the CIDEr metric is not able to correctly capture this trend, as shown by the results in Table~\ref{tab:cider-stab}, which indicate
a flat/reverse trend.

\subsection{Caption is Translation of Stabilizer}
So far, we have verified that both the target-language Caption and the Stabilizer English outputs for the \plugs-2L models are better compared
to the alternative ways of producing them.
Additionally, we want to check whether the Stabilizer and the target-language Caption are actually translations of each other, and not just
independently good captions associated with the input image.
In Table~\ref{tab:bleu}, we show the BLEU-4 score of the translation of the Stabilizer output for the \plugs-2L models, compared to the corresponding \plugs-2L Caption
treated as a reference, using the images in the \oid~test set.
The high BLEU scores are indeed confirming that the Caption outputs are close translations of the Stabilizer English outputs.
This allows us to conclude that \plugs~models are indeed performing the double-duty of captioning and translation.

\subsection{Stabilizers Used for Quality Estimation}
Finally, we perform an experiment to understand the extent to which
the quality of the Stabilizer outputs is correlated with the quality of the
target-language Captions, so that a QE model~\cite{levinboim2019quality-arxiv} trained for English
can be applied directly on \plugs~model outputs (more specifically, on the Stabilizer outputs).
To that end, we perform human evaluations of stand-alone captions.

% It looks like this figure does not bring much to the discussion at this point, and can be substituted by text-only.
% Given that we are over 8-pages with it included, I'm taking it out.
\begin{comment}
\subsubsection{Stand-alone Caption Evaluation}
\begin{figure*}[h!]
  \begin{center}
   \includegraphics[width=1.0\linewidth]{figs/standalone.png}
   \caption{\label{fig:standalone}: Standalone evaluation of one set of image captions.}
  \end{center}
\end{figure*}
\end{comment}

% As illustrated in Fig~\ref{fig:standalone},
In this type of evaluation, the raters are shown an image along with a single caption,
and are asked to provide an absolute rating for the caption on a 4-point scale.
As before, we define the metric $\ok$ = Percent of images where majority of raters (2 of 3) marked Caption as Acceptable, Good or Excellent.
Since these ratings are obtained individually for captions, we can use them to measure cross-lingual quality correlations.

\subsubsection{Quality Correlation between Stabilizer and Caption}
We use the stand-alone caption evaluation results to compute quality correlations.
Table~\ref{tab:corr} shows the correlation between the median human rating for
the Stabilizer (English caption) vs Caption (target-language caption) for the \plugs~models considered.
We see that the correlation is much higher compared to the baselines, calculated by computing
the correlation of the median rating for the Stabilizer vs Caption (target-language)
generated by the \tgt~and \ttg~approaches.

These results confirm that the \plugs~approach appears to be best suited for leveraging an existing En QE model, due to the availability of the
generated Stabilizer output that tends to maintain consistency between the English and the target-language caption, with respect to content accuracy.

\section{Conclusions}
\label{sec:conclusions}
We present a cross-modal language generation approach called \plugs,
which successfully combines the availability of an existing gold annotation (usually in English)
with the availability of translation engines that automatically produce silver-data annotations.
The result is a multilingual engine capable of generating high-quality outputs in the target languages, with no gold annotations needed for these languages.

We show that, for image captioning, the \plugs~approach out-performs other alternatives, while also providing the ability
to pack multiple languages in a single model for increased performance.
Surprisingly, by considering the generated outputs in the original language of the annotation (Stabilizer outputs),
we find that the quality of the Stabilizers is higher compared to the outputs of a model trained on the original annotated data.

Overall, our results can be understood as a successful instance of transfer learning from a uni-modal task (text-to-text translation) to a cross-modal task (image-to-text generation),
which allows us to indirectly leverage the abundance of text-only parallel data annotations across many languages to improve the quality
of an annotation-poor cross-modal setup.

\bibliography{garcon.bib}
\bibliographystyle{acl_natbib}

\end{document}